\documentclass[10pt,twocolumn,letterpaper]{article}

\usepackage{cvpr}
\usepackage{times}
\usepackage{epsfig}
\usepackage{graphicx}
\usepackage{amsmath}
\usepackage{amssymb}
\usepackage{caption,subcaption}

\usepackage[pagebackref=true,breaklinks=true,letterpaper=true,colorlinks,bookmarks=false]{hyperref}

\cvprfinalcopy 

\def\cvprPaperID{13} 

\ifcvprfinal\fi
\begin{document}

\title{Fine-grained Activity Recognition in Baseball Videos}

\author{AJ Piergiovanni and Michael S. Ryoo\\
Department of Computer Science, Indiana University, Bloomington, IN 47408 \\
\texttt{\{ajpiergi,mryoo\}@indiana.edu}\\
}

\maketitle
\thispagestyle{empty}
\global\csname @topnum\endcsname 0
\global\csname @botnum\endcsname 0

\begin{abstract}
In this paper, we introduce a challenging new dataset, MLB-YouTube, designed for fine-grained activity detection. The dataset contains two settings: segmented video classification as well as activity detection in continuous videos. We experimentally compare various recognition approaches capturing temporal structure in activity videos, by classifying segmented videos and extending those approaches to continuous videos. We also compare models on the extremely difficult task of predicting pitch speed and pitch type from broadcast baseball videos. We find that learning temporal structure is valuable for fine-grained activity recognition.
\end{abstract}

\section{Introduction}
Activity recognition is an important problem in computer vision with many applications within sports. Every major professional sporting event is recorded for entertainment purposes, but is also used for analysis by coaches, scouts, and media analysts. Many game statistics are currently manually tracked, but could be replaced by computer vision systems. Recently, the MLB has used the PITCHf/x and Statcast systems that are able to automatically capture pitch speed and motion. These systems use multiple high-speed cameras and radar to capture detailed measurements for every player on the field. However, much of this data is not publicly available.

\begin{figure}
    \centering
    \includegraphics[width=\linewidth]{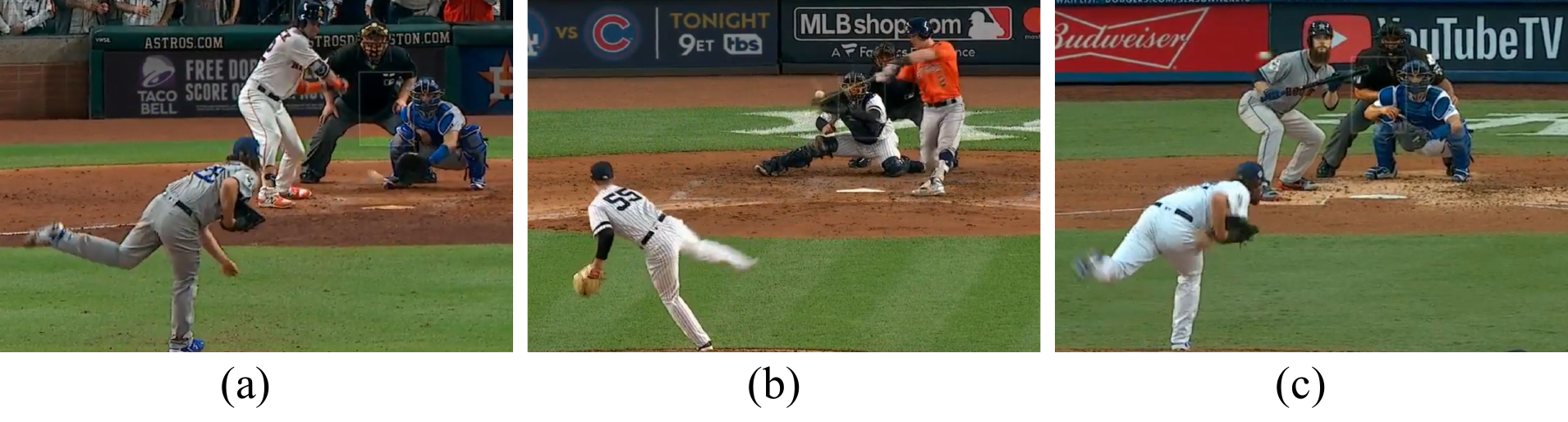}
    \caption{Examples of (a) No swing, (b) Swing and (c) Bunting. This task is quite challenging as the difference between these activities is very small.}
    \label{fig:intro}
\end{figure}

In this paper, we introduce a new dataset, MLB-YouTube, which contains densely annotated frames with activities from broadcast baseball videos. Unlike many existing activity recognition or detection datasets, ours focuses on fine-grained activity recognition. As shown in Fig.~\ref{fig:intro}, the scene structure is very similar between activities, often the only difference is the motion of a single person. Additionally, we only have a single camera viewpoint to determine the activity. We experimentally compare various approaches for temporal feature pooling for both segmented video classification as well as activity detection in continuous videos.

\section{Related Works}
Activity recognition has been a popular research topic in computer vision~\cite{aggarwal11,karpathy2014large,simonyan2014two,wang2011action,ryoo13}. Hand-crafted features, such as dense trajectories~\cite{wang2011action} gave promising results on many datasets. More recent works have focused on learning CNNs for activity recognition~\cite{carreira2017quo,tran2014c3d}. Two-stream CNNs take spatial RGB frames and optical flow frames as input~\cite{simonyan2014two,feichtenhofer2016convolutional}. 3D XYT convoltuional models have been trained to learn spatio-temporal features~\cite{tran2014c3d,carreira2017quo,tran2017convnet,hara2017learning}. To train these CNN models, large scale datasets such as Kinetics~\cite{kay2017kinetics}, THUMOS~\cite{THUMOS14}, and ActivityNet~\cite{caba2015activitynet} have been created.

Many works have explored temporal feature aggregation for activity recognition. Ng et al.~\cite{ng2015beyond} compared various pooling methods and found that LSTMs and max-pooling the entire video performed best. Ryoo et al.~\cite{ryoo2015pooled} found that pooling intervals of different locations/lengths was beneficial to activity recognition. Piergiovanni et al.~\cite{piergiovanni2017learning} found that learning important sub-event intervals and using those for classification improved performance.

Recently, segment-based 3D CNNs have been used to capture spatio-temporal information simultaneously for activity detection~\cite{xu2017r,shou2016temporal,shou2017cdc}.  These approaches all rely on the 3D CNN to capture temporal dynamics, which usually only contain 16 frames. Some works have studied longer-term temporal structures ~\cite{carreira2017quo,karpathy2014large,ng2015beyond,varol17}, but it was generally done with a temporal pooling of local representations or (spatio-)temporal convolutions with larger fixed intervals. Recurrent neural networks (RNNs) also have been used to model activity transitions between frames~\cite{yeung2015every,yeung2016end,escorcia2016daps}.
\begin{figure*}
    \centering
    \includegraphics[width=\textwidth]{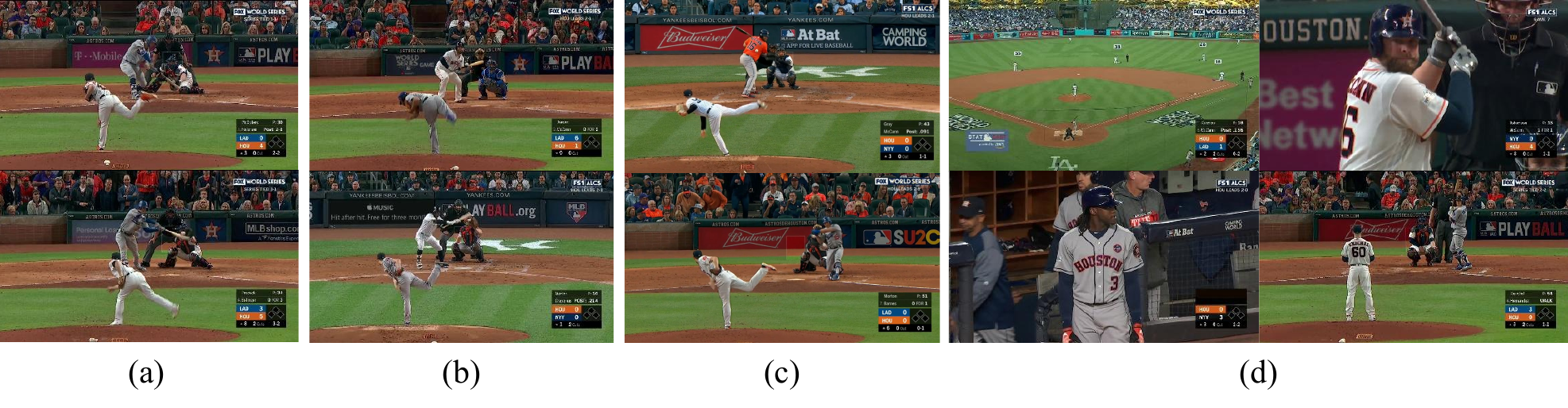}
    \caption{Examples of some of the activities in the MLB-YouTube Dataset. The activities are (a) Hit, (b) Bunt, (c) Hit by pitch, and (d) No activity (hard negatives). The difference between the activities is quite small, making this a challenging task.}
    \label{fig:activity-examples}
\end{figure*}

\section{MLB-YouTube Dataset}

We created a large-scale dataset consisting of 20 baseball games from the 2017 MLB post-season available on YouTube with over 42 hours of video footage. Our dataset consists of two components: segmented videos for activity recognition and continuous videos for activity classification. Our dataset is quite challenging as it is created from TV broadcast baseball games where multiple different activities share the camera angle. Further, the motion/appearance difference between the various activities is quite small (e.g., the difference between swinging the bat and bunting is very small), as shown in Fig.~\ref{fig:activity-examples}. Many existing activity detection datasets, such as THUMOS~\cite{THUMOS14} and ActivityNet~\cite{caba2015activitynet}, contain a large variety of activities that vary in setting, scale, and camera angle. This makes even a single frame from one activity (e.g., swimming) to be very different from that of another activity (e.g., basketball). On the other hand, a single frame from one of our baseball videos is often not enough to classify the activity.

\begin{figure}
    \centering
    \includegraphics[width=\linewidth]{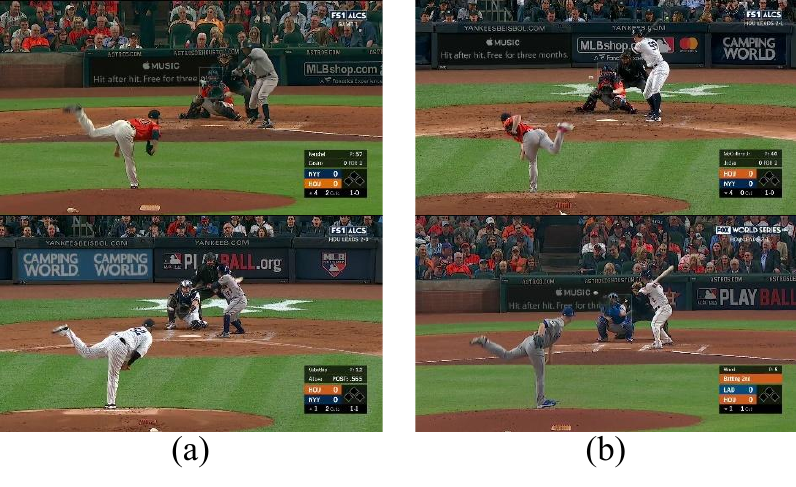}
    \caption{The difference between a (a) strike and (b) ball is very small.}
    \label{fig:ex-ball-strike}
\end{figure}

\begin{figure}
    \centering
    \includegraphics[width=0.8\linewidth]{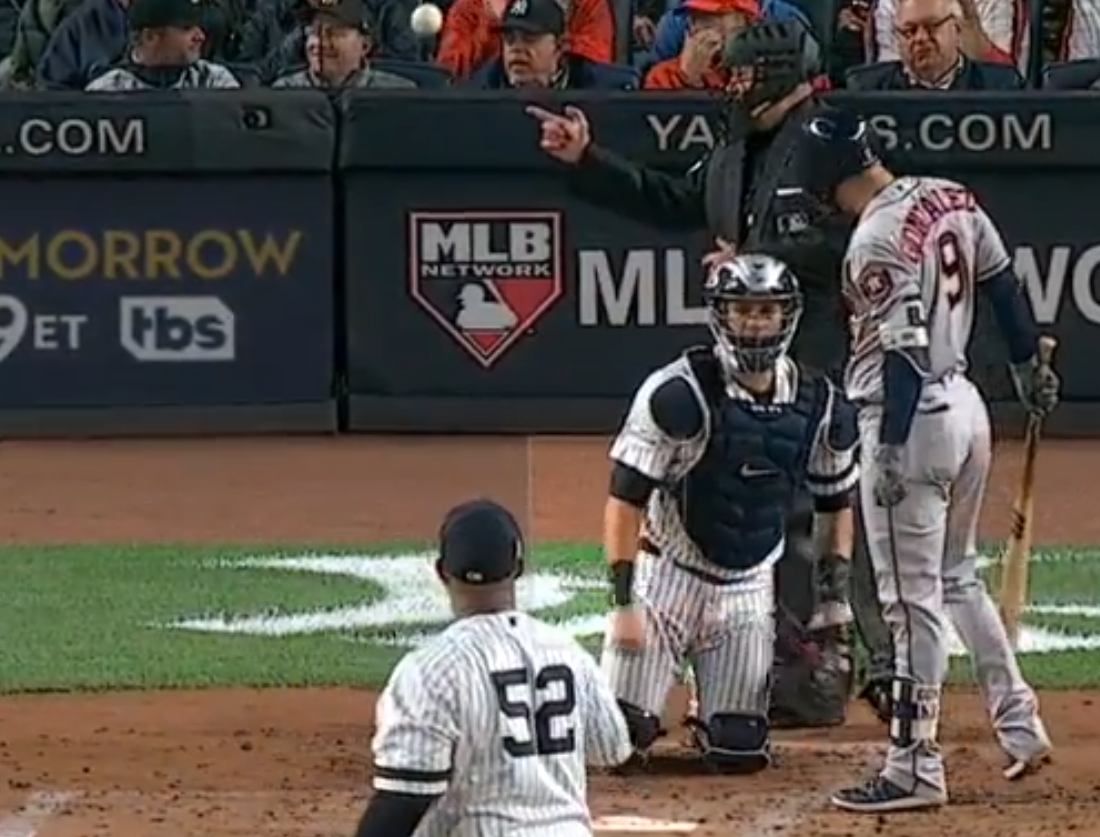}
    \caption{Detecting a strike often relies on the umpire's signal (or lack of signal) at the end of the pitch sequence.}
    \label{fig:signal}
\end{figure}

Fig.~\ref{fig:ex-ball-strike} shows the small difference between a ball and strike. To distinguish these activities requires detecting if the batter swings or not, or detecting the umpire's signal (Fig.~\ref{fig:signal}) for a strike, or no signal for a ball. Further complicating this task is that the umpire can be occluded by the batter or catcher and each umpire has a unique way to signal a strike.

Our segmented video dataset consists of 4,290 video clips. Each clip is annotated with the various baseball activities that occur, such as swing, hit, ball, strike, foul, etc. A video clip can contain multiple activities, so we treat this as a multi-label classification task. A full list of the activities and the number of examples of each is shown in Table~\ref{tab:counts}. We additionally annotated each clip containing a pitch with the pitch type (e.g., fastball, curveball, slider, etc.) and the speed of the pitch. We also collected a set of 2,983 hard negative examples where no action occurs. These examples include views of the crowd, the field, or the players standing before or after a pitch occurred. Examples of the activities and hard negatives are shown in Fig.~\ref{fig:activity-examples}.

Our continuous video dataset consists of 2,128 1-2 minute long clips from the videos. Each video frame is annotated with the baseball activities that occur.  Each continuous clip contains on average of 7.2 activities, resulting in a total of over 15,000 activity instances. Our dataset and models are avaiable at \href{https://github.com/piergiaj/mlb-youtube/}{https://github.com/piergiaj/mlb-youtube/}


\begin{table}
\caption{The activity classes in the segmented MLB-YouTube dataset and the number of instances of the activity.}
\label{tab:counts}
\centering
\begin{tabular}{c|c}
\hline
Activity & \# Examples \\
\hline
No Activity & 2983 \\
Ball     &  1434 \\
Strike   &  1799 \\
Swing    &  2506 \\
Hit      &  1391 \\
Foul     &  718 \\
In Play  &  679 \\
Bunt     &  24 \\
Hit by Pitch & 14\\
\hline
\end{tabular}
\end{table}

\section{Segmented Video Recognition Approach}
\begin{figure*}
    \centering
    \includegraphics[width=\textwidth]{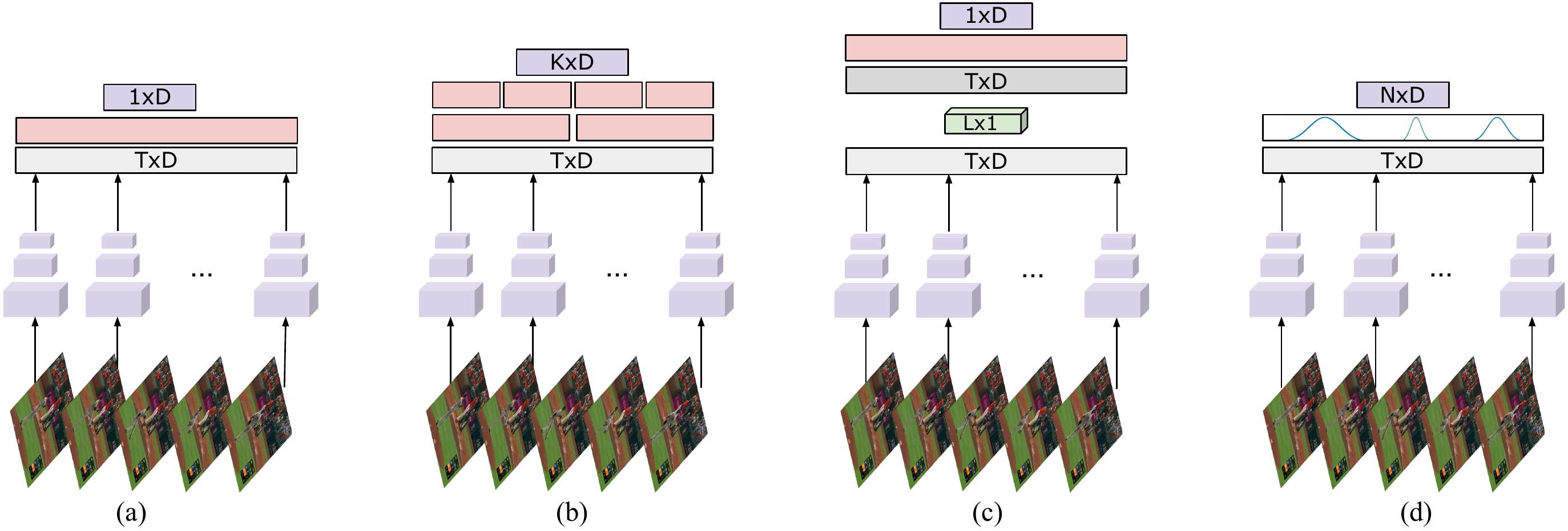}
    \caption{Illustration of the various feature aggregation methods. (a) Temporal max/mean-pooling, (b) Temporal Pyramid Pooling, (c) Temporal convolution followed by temporal max-pooling, (d) Sub-events. All models are followed by a fully-connected layer for classification.}
    \label{fig:segment-models}
\end{figure*}

We explore various methods of temporal feature aggregation for segmented video activity recognition. With segmented videos, the classification task is much easier as every frame (in the video) corresponds to the activity. The model does not need to determine when an activity begins and ends. The base component of our approaches is based on a CNN providing a per-frame (or per-segment) representation. We obtain this from standard two-stream CNNs~\cite{simonyan2014two,feichtenhofer2016convolutional} using a recent deep CNNs such as I3D~\cite{carreira2017quo} or InceptionV3~\cite{szegedy2016rethinking}. 

Given $v$, the $T\times D$ features from a video, where $T$ is the temporal length of the video and $D$ is the dimensionality of the feature, the standard method for feature pooling is max- or mean-pooling over the temporal dimension followed by a fully-connected layer to classify the video clip~\cite{ng2015beyond}, as shown in Fig.~\ref{fig:segment-models}(a). However, this provides only one representation for the entire video, and loses valuable temporal information. One way to address this is to use a fixed temporal pyramid of various lengths, as shown in Fig~\ref{fig:segment-models}(b). We divide the input video into intervals of various lengths (1/2, 1/4, and 1/8), and max-pool each interval. We concatenate these pooled features together, resulting in a $K\times D$ representation ($K$ is the number of intervals in the temporal pyramid), and use a fully-connected layer to classify the clip. 

We also try learning temporal convolution filters, which can learn to aggregate local temporal structure. The kernel size is $L\times 1$ and it is applied to each frame. This allows each timestep representation to contain information from nearby frames. We then apply max-pooling over the output of the temporal convolution and use a fully-connected layer to classify the clip, shown in Fig.~\ref{fig:segment-models}(c). 

While temporal pyramid pooling allows some structure to be preserved, the intervals are predetermined and fixed. Previous works have found learning the sub-interval to pool was beneficial to activity recognition~\cite{piergiovanni2017learning}. The learned intervals are controlled by 3 learned parameters, a center $g$, a width  $\sigma$ and a stride $\delta$ used to parameterize $N$ Gaussians.  Given $T$, the length of the video, we first compute the locations of the strided Gaussians as:
\begin{equation} \label{eq:mrf_match}
\begin{split}
g_n &= 0.5\cdot T \cdot (\widetilde{g}_n + 1) \\
\delta_n &= \frac{T}{N-1} \widetilde{\delta}_n \\
\mu_n^i &= g_n + (i - 0.5N +0.5)\delta_n \\
\end{split}
\end{equation}

The filters are then created as:
\begin{equation} \label{eq:filters}
\begin{split}
F_m[i,t] &= \frac{1}{Z_{m}} \exp(-\frac{(t-\mu_m^i)^2}{2\sigma_m^2}) \\
& i\in\{0,1,\ldots,N-1\},~ t\in\{0,1,\ldots,T-1\} \\
\end{split}
\end{equation}
where $Z_m$ is a normalization constant.

We apply $F$ to the $T\times D$ video representation by matrix multiplication, resulting in a $N\times D$ representation which is used as input to a fully connected layer for classification. This method is shown in Fig~\ref{fig:segment-models}(d).

Other works have used LSTMs~\cite{ng2015beyond,donahue2015long} to model temporal structure in videos. We also compare to a bi-directional LSTM with 512 hidden units where we use the last hidden state as input to a fully-connected layer for classification.

We formulate our tasks as multi-label classification and train these models to minimize binary cross entropy:
\begin{equation}
    L(v) = \sum_c z_c\log(p(c|G(v))) + (1-z_c)\log(1-p(c|G(v)))
\end{equation}
Where $G(v)$ is the function that pools the temporal information (i.e., max-pooling, LSTM, temporal convolution, etc.), and $z_c$ is the ground truth label for class $c$.

\section{Activity Detection in Continuous Videos}
Activity detection in continuous videos is a more challenging problem. Here, our objective is to classify each frame with the occurring activities. Unlike segmented videos, there are multiple instances of activities occurring sequentially, often separated by frames with no activity. This requires the model to learn to detect the start and end of activities. As a baseline, we train a single fully-connected layer as a per-frame classifier. This method uses no temporal information not present in the features.

We extend the approaches presented for segmented video classification to continuous videos by applying each approach in a temporal sliding window fashion. To do this, we first pick a fixed window duration (i.e., a temporal window of $L$ features). We apply max-pooling to each window (as in Fig.~\ref{fig:segment-models}(a)) and classify each pooled segment.

We can similarly extend temporal pyramid pooling. Here, we split the window of length $L$ into segments of length $L/2, L/4, L/8$, this results in 14 segments for each window. We apply max-pooling to each segment and concatenate the pooled features together. This gives a $14\times D$-dim representation for each window which is used as input to the classifier.

For temporal convolutional models on continuous videos, we slightly alter the segmented video approach. Here, we learn a temporal convolutional kernel of length $L$ and convolve it with the input video features. This operation takes input of size $T\times D$ and produces output of size $T\times D$. We then apply a per-frame classifier on this representation. This allows the model to learn to aggregate local temporal information.

To extend the sub-event model to continuous videos, we follow the approach above, but set $T=L$ in Eq.~\ref{eq:mrf_match}. This results in filters of length $L$. Given $v$, the $T\times D$ video representation, we convolve (instead of using matrix multiplication) the sub-event filters, $F$, with the input, resulting in a $N\times D\times T$-dim representation. We use this as input to a fully-connected layer to classify each frame.

We train the model to minimize the per-frame binary classification:
\begin{equation}
\begin{split}
    L(v) = \sum_{t,c} &z_{t,c}\log(p(c|H(v_t))) + \\
    &(1-z_{t,c})\log(1-p(c|H(v_t)))    
\end{split}
\end{equation}
where $v_t$ is the per-frame or per-segment feature at time $t$, $H(v_t)$ is the sliding window application of one of the feature pooling methods, and $z_{t,c}$ is the ground truth class at time $t$.

A recent approach to learn `super-events' (i.e., global video context) was proposed and found to be effective for activity detection in continuous videos~\cite{piergiovanni2018super}. The approach learns a set of temporal structure filters that are modeled as a set of $N$ Cauchy distributions. Each distribution learns a center, $x_n$ and a width, $\gamma_n$. Given $T$, the length of the video, the filters are constructed by:
\begin{equation}
\begin{split}
  \hat{x_n} &= \frac{(T - 1) \cdot (\tanh\left(x_n\right)+1)}{2}\\
  \hat{\gamma_n} &= \exp(1 - 2\cdot|\tanh\left(\gamma_n\right)|)\\
  F[t,n] &= \cfrac{1}{Z_n\pi\hat{\gamma_n}\left(\cfrac{(t-\hat{x_n})}{\hat{\gamma_n}}\right)^2}
\end{split}
\end{equation}
where $Z_n$ is a normalization constant, $t\in\{1,2,\ldots, T\}$ and $n\in\{1,2,\ldots,N\}$.

The filters are combined with learned per-class soft-attention weights, $A$ and the super-event representation is computed as:
\begin{equation}
    S_c = \sum_m^{M} A_{c,m}\cdot \sum_t^T F_{m}[t] \cdot v_t
\end{equation}
where $v$ is the $T\times D$ video representation. These filters allow the model to learn intervals to focus on for useful temporal context. The super-event representation is concatenated to each timestep and used for classification. We also try concatenating the super- and sub-event representations to use for classification to create a three-level hierarchy of event representation.

\section{Experiments}
\subsection{Implementation Details}

As our base per-segment CNN, we use the I3D~\cite{carreira2017quo} network pretrained on the ImageNet and Kinetics~\cite{kay2017kinetics} datasets. I3D obtained state-of-the-art results on segmented video tasks, and this allows us to obtain reliable per-segment feature representation. We also use two-stream version of InceptionV3~\cite{szegedy2016rethinking} pretrained on Imagenet and Kinetics as our base per-frame CNN, and compared them. We chose InceptionV3 as it is deeper than previous two-stream CNNs such as \cite{simonyan2014two,feichtenhofer2016convolutional}. We extracted frames from the videos at 25 fps, computed TVL1~\cite{zach2007duality} optical flow, clipped to $[-20,20]$. For InceptionV3, we computed features for every 3 frames (8 fps). For I3D, every frame was used as the input. I3D has a temporal stride of 8, resulting in 3 features per second (3 fps). We implemented the models in PyTorch. We trained our models using the Adam~\cite{kingma2014adam} optimizer with the learning rate set to 0.01. We decayed the learning rate by a factor of 0.1 after every 10 training epochs. We trained our models for 50 epochs. Our source code, dataset and trained models are available at \href{https://github.com/piergiaj/mlb-youtube/}{https://github.com/piergiaj/mlb-youtube/}

\subsection{Segmented Video Activity Recognition}

We first performed the binary pitch/non-pitch classification of each video segment. This task is relatively easy, as the difference between pitch frames and non-pitch frames are quite different. The results, shown in Table~\ref{tab:segmented-binary-pitch}, do not show much difference between the various features or models.

\begin{table}
\caption{Results on segmented videos performing binary pitch/non-pitch classification.}
\label{tab:segmented-binary-pitch}
\centering
\begin{tabular}{c|ccc}
\hline
Model                     & RGB   & Flow  & Two-stream \\
\hline
InceptionV3               & 97.46 & 98.44 & 98.67 \\
InceptionV3 + sub-events  & 98.67 & 98.73 & \bf{99.36} \\
I3D                       & 98.64 & 98.88 & 98.70 \\
I3D + sub-events          & 98.42 & 98.35 & 98.65 \\
\hline
\end{tabular}
\end{table}

\subsubsection{Multi-label Classification}
We evaluate and compare the various approaches of temporal feature aggregation by computing mean average precision (mAP) for each video clip, which is a standard evaluation metric for multi-label classification tasks. Table~\ref{tab:segmented-compare} compares the performance of the various temporal feature pooling methods. We find that all approaches outperform mean/max-pooling, confirming that maintaining some temporal structure is important for activity recognition. We find that fixed temporal pyramid pooling and LSTMs give some improvement. Temporal convolution provides a larger increase in performance, however it requires significantly more parameters (see Table~\ref{tab:efficiency}). Learning sub-events of~\cite{piergiovanni2017learning} we found to give the best performance on this task. While LSTMs and temporal have been previously used for this task, they require greater number of parameters and perform worse, likely due to overfitting. Additionally, LSTMs require the video features to be processes sequentially as each timestep requires the output from the previous timestep, while the other approaches can be completely parallelized.

\begin{table}
\caption{Additional number of parameters for models when added to base (e.g., I3D or Inception V3).}
\label{tab:efficiency}
\centering
\begin{tabular}{c|c}
\hline
Model & \# Parameters \\
\hline
Max/Mean Pooling  &  16K  \\
Pyramid Pooling   &  115K \\
LSTM              & 10.5M \\
Temporal Conv     & 31.5M \\
Sub-events        &  36K \\
\hline
\end{tabular}
\end{table}

\begin{table}
\caption{mAP results on segmented videos performing multi-label classification. We find that learning sub-intervals to pool is important for activity recognition.}
\label{tab:segmented-compare}
\centering
\small
\begin{tabular}{c|ccc}
\hline
Method                                & RGB   & Flow  & Two-stream \\
\hline
Random                               & 16.3 & 16.3 & 16.3 \\
\hline
InceptionV3 + mean-pool             & 35.6 & 47.2 & 45.3 \\
InceptionV3 + max-pool              & 47.9 & 48.6 & 54.4 \\
InceptionV3 + pyramid               & 49.7 & 53.2 & 55.3 \\
InceptionV3 + LSTM                  & 47.6 & 55.6 & 57.7 \\
InceptionV3 + temporal conv         & 47.2 & 55.2 & 56.1 \\
InceptionV3 + sub-events            & 56.2 & 62.5 & \bf{62.6} \\
\hline
I3D + mean-pool                    & 42.4 & 47.6 & 52.7 \\
I3D + max-pool                     & 48.3 & 53.4 & 57.2 \\
I3D + pyramid                      & 53.2 & 56.7 & 58.7 \\
I3D + LSTM                         & 48.2 & 53.1 & 53.1 \\
I3D + temporal conv                & 52.8 & 57.1 & 58.4 \\
I3D + sub-events                   & 55.5 & 61.2 & 61.3 \\
\hline
\end{tabular}
\end{table}

\begin{figure*}
    \centering
      \includegraphics[width=\linewidth]{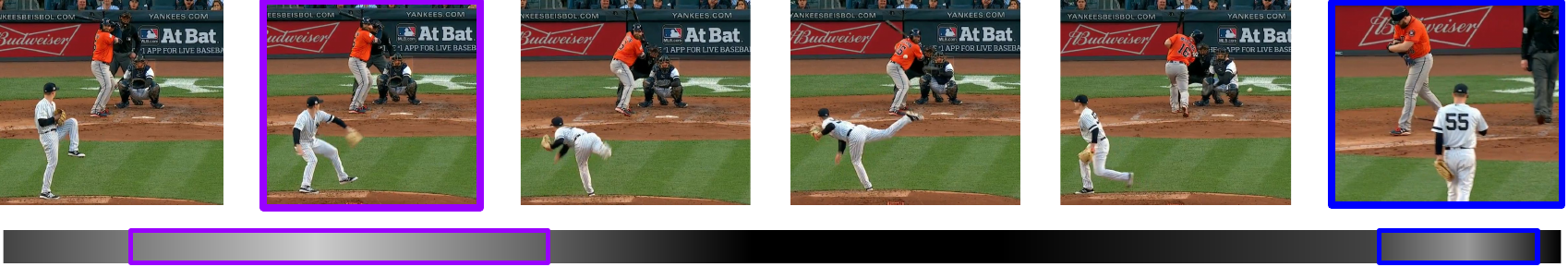}
      \caption{Illustration of the learned sub-events for the hit by pitch activity. It localizes the start of the pitch as well as the batter walking towards first base.}
      \label{fig:sub-event-examples-hbp}
\end{figure*}
\begin{figure*}
        \centering
        \includegraphics[width=\linewidth]{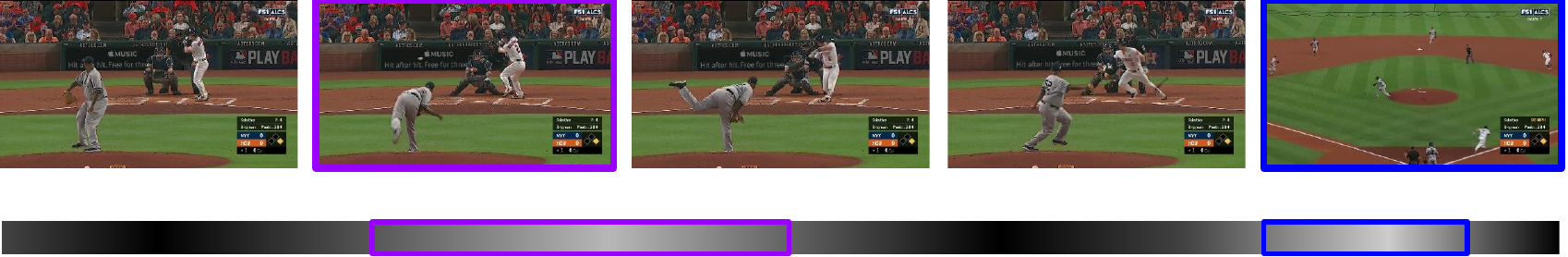}
        \caption{Illustration of the learned sub-event for the hit activity. It captures the start of the pitch as well as the camera change to watch the runner and ball in the field.}
        \label{fig:sub-event-examples-hit}
\end{figure*}

In Table~\ref{tab:segmented-multilabel}, we compare the average precision for each activity class. Learning temporal structure is especially helpful for frame-based features (e.g., InceptionV3) whose features capture minimal temporal information when compared to segment-based features (e.g., I3D) which capture some temporal information. Additionally, we find that sub-event learning helps especially in the case of strikes, hits, foul balls, and hit by pitch, as those all have changes in video features after the event. For example, after the ball is hit, the camera will often follow the ball's trajectory, while being hit by a pitch the camera will follow the player walking to first base, as shown in Fig.~\ref{fig:sub-event-examples-hbp} and Fig.~\ref{fig:sub-event-examples-hit}.

\begin{table*}
\caption{Per-class average precision for segmented videos performing multi-label activity classification using two-stream features. We find that using sub-events to learn temporal intervals of interest is beneficial to activity recognition.}
\label{tab:segmented-multilabel}
\centering
\begin{tabular}{c|cccccccc}
\hline
Method                     & Ball   & Strike   &  Swing  &  Hit   &  Foul  &  In Play  &  Bunt  &  Hit by Pitch  \\
\hline
Random                    & 21.8   &  28.6    &   37.4  &  20.9  &  11.4  &  10.3     &  1.1   & 4.5   \\
\hline
InceptionV3 + max-pool    & 60.2   &  84.7    &   85.9  &  80.8  &  40.3  &  74.2     &  10.2   & 15.7  \\
InceptionV3 + sub-events  & 66.9   &  93.9    &   90.3  &  90.9  &  60.7  &  89.7     &  12.4   & 29.2  \\
I3D  + max-pool           & 59.4   &  90.3    &   87.7  &  85.9  &  48.1  &  76.1     &  14.3   & 18.2  \\
I3D + sub-events          & 62.5   &  91.3    &   88.5  &  86.5  &  47.3  &  75.9     &  16.2   & 21.0  \\
\hline
\end{tabular}
\end{table*}

\subsubsection{Pitch Speed Regression}
\begin{figure*}
    \centering
    \includegraphics[width=\linewidth]{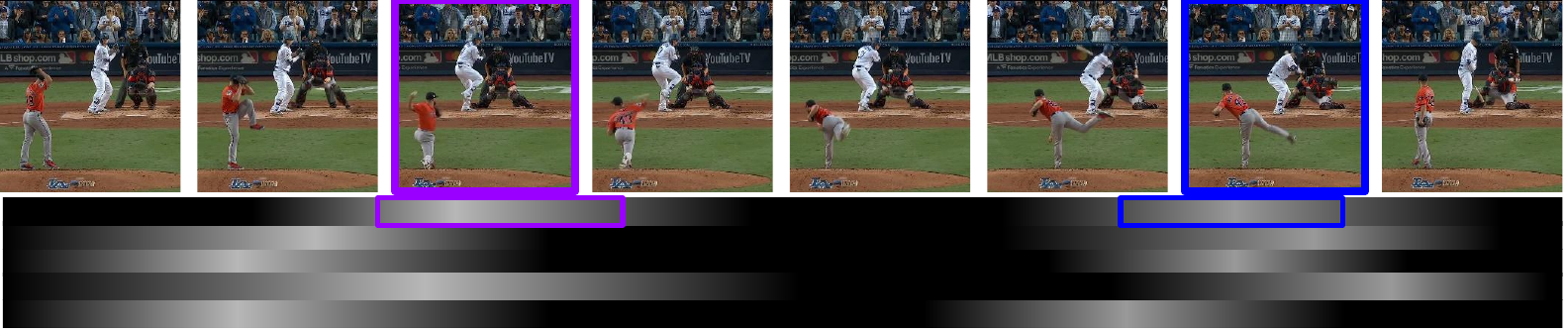}
    \caption{Illustration of the learned sub-events used for the pitch speed regression task. Each sub-event localizes different start/end times for the pitch and this information is used to predict the pitch speed.}
    \label{fig:pitch-speed-ex}
\end{figure*}

Pitch speed regression from video frames is a challenging task because it requires the network to learn localize the start of a pitch and the end of the pitch, then compute the speed from a weak signal (i.e., only pitch speed). The baseball is often small and occluded by the pitcher. Professional baseball pitchers can throw the ball in excess of 100mph and the pitch only travels 60.5 ft. Thus the ball is only traveling for roughly 0.5 seconds. Using our initial frame rates of 8fps and 3fps, there was only 1-2 features of the pitch in the air, which we found was not enough to determine pitch speed. The YouTube videos contain 60fps, so we recomputed optical flow and extract RGB frames at 60fps. We use a fully-connected layer with one output to predict the pitch speed and minimize the $L_1$ loss between the ground truth speed and predicted speed. Using features extracted at 60fps, we were able to determine pitch speed, with 3.6mph average error. Table~\ref{tab:speed-regression} compares various models. Fig.~\ref{fig:pitch-speed-ex} shows the sub-event learned for various speeds. 

\begin{table}
\caption{Results on segmented video pitch speed regression. We are reporting their root-mean-squared errors.}
\label{tab:speed-regression}
\centering
\begin{tabular}{c|c}
\hline
Method                     & Two-stream \\
\hline
I3D                       & 4.3 mph  \\
I3D + LSTM                & 4.1 mph  \\
I3D + sub-events          & 3.9 mph  \\
InceptionV3               & 5.3 mph  \\
InceptionV3 + LSTM        & 4.5 mph  \\
InceptionV3 + sub-events  & \bf{3.6} mph  \\
\hline
\end{tabular}
\end{table}

\subsubsection{Pitch Type Classification}
We experiment to see if it is possible to predict the pitch type from video. This is an extremely challenging problem because it is adversarial; pitchers practice to disguise their pitch from batters. Additionally, the difference between pitches can be as small as a difference in grip on the ball and which way it rotates with respect to the laces, which is rarely visible in broadcast baseball videos.  In addition to the video features used in the previous experiments, we also extract pose using OpenPose~\cite{cao2017realtime}. Our features are heatmaps of joint and body part locations which we stack along the channel axis and use as input to an InceptionV3 CNN which we newly train on this task. We chose to try pose features as the body mechanics can vary between pitches as well (e.g., the stride length and arm angles can be different for fastballs and curveballs). Our dataset has 6 different pitches (fastball, sinker, curveball, changeup, slider, and knuckle-curve). We report our results in Table~\ref{tab:pitch-type}. We find that LSTMs actually perform worse than the baseline, likely due to overfitting the small differences between pitch types, while learning sub-events helps. We observe that fastballs are the easiest to detect (68\% accuracy) followed by sliders (45\% accuracy), while sinkers are the hardest to classify (12\%).

\begin{table}
\caption{Accuracy of our pitch type classification using I3D on video inputs and InceptionV3 trained using pose heatmaps as input.}
\label{tab:pitch-type}
\centering
\begin{tabular}{c|c}
\hline
Method                    & Accuracy  \\
\hline
Random                    & 17.0\%  \\
\hline
I3D                       & 25.8\%  \\
I3D + LSTM                & 18.5\% \\
I3D + sub-events          & 34.5\% \\
\hline
Pose                      & 28.4\%  \\
Pose + LSTM               & 27.6\%  \\
Pose + sub-events         & \bf{36.4}\%  \\
\hline
\end{tabular}
\end{table}

\subsection{Continuous Video Activity Detection}
We evaluate the extended models on continuous videos using per-frame mean average precision (mAP), and the results are shown in Table~\ref{tab:continuous-classify}. This setting is more challenging that the segmented videos as the model must determine when the activity starts and ends and contains negative examples that are more ambiguous than the hard negatives in the segmented dataset (e.g., the model has to determine when the pitch event begins compared to just the pitcher standing on the mound).  We find that all models improve over the baseline per-frame classification, confirming that temporal information is important for detection. We find that fixed temporal pyramid pooling outperforms max-pooling. The LSTM and temporal convolution seem to overfit, due to the larger number of parameters. We find that the convolutional form of sub-events to pool local temporal structure especially helps frame based features, not as much on segment features. Using the super-event approach~\cite{piergiovanni2018super}, further improves performance. Combining the convolutional sub-event representation with the super-event representation provides the best performance.

\begin{table}
\caption{Results on continuous videos performing multi-label activity classification (per-frame mAP).}
\label{tab:continuous-classify}
\centering
\small
\begin{tabular}{c|ccc}
\hline
Method                            & RGB   & Flow  & Two-stream \\
\hline
Random                           & 13.4  & 13.4 & 13.4 \\
\hline
I3D                              & 33.8 & 35.1 & 34.2 \\
I3D + max-pooling                & 34.9 & 36.4 & 36.8 \\
I3D + pyramid                    & 36.8 & 37.5 & 39.7 \\
I3D + LSTM                       & 36.2 & 37.3 & 39.4 \\
I3D + temporal conv              & 35.2 & 38.1 & 39.2 \\
I3D + sub-events                 & 35.5 & 37.5 & 38.5 \\
I3D + super-events               & 38.7 & 38.6 & 39.1 \\
I3D + sub+super-events           & 38.2 & 39.4 & 40.4 \\
\hline
InceptionV3                      & 31.2 & 31.8 & 31.9 \\
InceptionV3 + max-pooling        & 31.8 & 34.1 & 35.2 \\
InceptionV3 + pyramid            & 32.2 & 35.1 & 36.8 \\
InceptionV3 + LSTM               & 32.1 & 33.5 & 34.1 \\
InceptionV3 + temporal conv      & 28.4 & 34.4 & 33.4 \\
InceptionV3 + sub-events         & 32.1 & 35.8 & 37.3 \\
InceptionV3 + super-events       & 31.5 & 36.2 & 39.6 \\
InceptionV3 + sub+super-events   & 34.2 & 40.2 & \bf{40.9} \\
\hline
\end{tabular}
\end{table}

\section{Conclusion}
We introduced a challenging new dataset, MLB-YouTube, for fine-grained activity recognition in videos. We experimentally compare various recognition approaches with temporal feature pooling for both segmented and continuous videos. We find that learning sub-events to select the temporal regions-of-interest provides the best performance for segmented video classification. For detection in continuous videos, we find that learning convolutional sub-events combined with the super-event representation to form a three-level activity hierarchy provides the best performance.


{\small
\bibliographystyle{ieee}
\bibliography{bib}
}

\end{document}